\documentclass[journal,twoside,web]{ieeecolor}
\usepackage{jsen}
\usepackage{cite}
\usepackage{amsmath,amssymb,amsfonts}
\usepackage{algorithmic}
\usepackage{graphicx}
\usepackage{textcomp}
\usepackage{wrapfig}

\usepackage{etoolbox}
\makeatletter
\@ifundefined{color@begingroup}%
  {\let\color@begingroup\relax
   \let\color@endgroup\relax}{}%
\def\fix@ieeecolor@hbox#1{%
  \hbox{\color@begingroup#1\color@endgroup}}
\patchcmd\@makecaption{\hbox}{\fix@ieeecolor@hbox}{}{}
\patchcmd\@makecaption{\hbox}{\fix@ieeecolor@hbox}{}{}
\makeatother

\def\BibTeX{{\rm B\kern-.05em{\sc i\kern-.025em b}\kern-.08em
    T\kern-.1667em\lower.7ex\hbox{E}\kern-.125emX}}
\definecolor{abstractbg}{rgb}{0.89804,0.94510,0.83137}
\begin{document}
\title{Component-Aware Structure-Preserving Style Transfer for Satellite Visual Sim2Real Data Construction}
\author{Zongwu Xie, Yonglong Zhang, Yifan Yang, Yang Liu\textsuperscript{\(\dagger\)}, and Baoshi Cao ,Guanghu Xie
\thanks{\textsuperscript{\(\dagger\)}Corresponding author: Yang Liu (e-mail: liuyanghit@hit.edu.cn).}
\thanks{\textsuperscript{*}This work was supported by the Natural Science Foundation of Heilongjiang Province for Excellent Young Scholars (Grant No. YQ2024E018).}
\thanks{All authors are with the State Key Laboratory of Robotics and Systems, Harbin Institute of Technology, Harbin 150001, Heilongjiang, China.}}

\IEEEtitleabstractindextext{%
\fcolorbox{abstractbg}{abstractbg}{%
\begin{minipage}{\textwidth}%
\begin{wrapfigure}[15]{r}{3in}%
\includegraphics[width=2.8in]{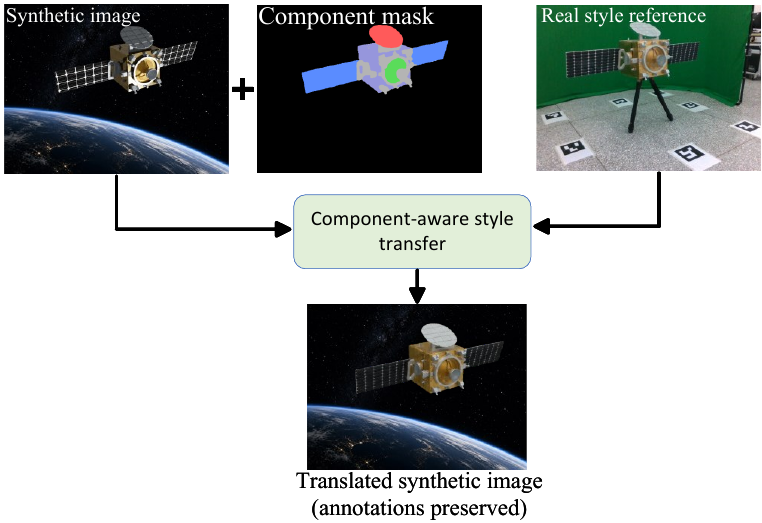}%
\end{wrapfigure}%
\begin{abstract}
For camera-based satellite visual sensing, Sim2Real data construction requires images that approach real-domain sensor appearance while retaining the annotations inherited from simulation. Real sensor images of satellite targets with reliable pose labels and component-level masks are difficult to acquire at scale, whereas synthetic rendering provides exact geometric annotations but suffers from a visible appearance gap. This paper presents a component-aware structure-preserving style transfer framework for satellite visual synthetic-to-real data construction. The method builds weakly paired real--synthetic samples from calibrated real acquisition, ArUco-based camera-pose measurement, CAD rendering, and component masks. It then extracts part-wise real-domain style codes from unlabeled real images and injects them into corresponding synthetic satellite regions through mask-aligned modulation. To keep the generated images usable for downstream sensor-data supervision, adversarial training is combined with local contrastive consistency, self-regularization, and edge-preserving constraints. Experiments are conducted on 5,000 rendered satellite images and 100 real images captured in a calibrated setup. The real images provide target-domain appearance references and final evaluation images, while the downstream GDRNet pose estimator is trained only on synthetic or translated synthetic images. Compared with representative image-translation baselines, the proposed method achieves the lowest image distribution discrepancy, with an FID of 54.32 and a KID of 0.048. When the translated data are used to train GDRNet in this target-domain adaptation setting, the ADD pass rate improves to 0.260 and the AUC improves to 0.611. These results indicate that component-level appearance transfer can improve annotation-preserving satellite visual Sim2Real data generation in the considered calibrated setup.
\end{abstract}

\begin{IEEEkeywords}
Satellite visual sensing, sensor data processing, Sim2Real data construction, style transfer, synthetic-to-real transfer, component-aware image translation.
\end{IEEEkeywords}
\end{minipage}}}
\maketitle

\section{Introduction}
\label{sec:introduction}

Reliable camera-based visual sensing of non-cooperative satellites is a prerequisite for on-orbit servicing, autonomous rendezvous, proximity operations, and space situational awareness. In this setting, learning-based satellite sensor-data processing systems require training images that cover viewpoint changes, illumination variation, target appearance, and background conditions. Reviews of cooperative and uncooperative spacecraft relative navigation, together with recent spaceborne pose benchmarks, show that image-based methods remain sensitive to these factors and to the synthetic-to-real gap~\cite{opromolla2017review,song2022relativeNavigationSurvey,pauly2023surveySpacecraftPose,sharma2020spn,kisantal2020speedChallenge,park2022speedPlus}.

The central difficulty is not only pose estimation itself, but also the data required to train it. Deep pose estimators generally need large numbers of images with accurate object poses, silhouettes, bounding boxes, and sometimes intermediate geometric cues. For satellite targets, obtaining such labels from real observations is difficult: the target is non-cooperative, the imaging geometry is hard to repeat, and component-level labels for solar panels, metallic bodies, antennas, docking rings, and nozzles require additional manual or semi-automatic annotation. Public spacecraft pose datasets and competitions have made this issue visible by combining synthetic images with limited real or hardware-in-the-loop data, while still emphasizing that the domain gap remains a major failure mode~\cite{kisantal2020speedChallenge,park2022speedPlus,park2024spnv2}.

CAD-based rendering offers an attractive way around the annotation bottleneck. Once a satellite model and camera configuration are available, a renderer can produce large-scale images together with exact poses, object masks, bounding boxes, and component masks. This strategy is common in object viewpoint and pose-estimation research, and it is also used in spacecraft-specific dataset generation and photorealistic spacecraft pose rendering~\cite{su2015renderForCNN,bechini2023spacecraftDataset,proenca2020photorealistic}. Yet rendered images do not automatically match real satellite observations. Material reflectance, solar-panel texture, specular response, camera noise, illumination, and background statistics can differ substantially between simulation and acquisition, so a model trained only on synthetic data may learn appearance cues that are unreliable in the real domain.

\begin{figure}[!t]
    \centering
    \includegraphics[width=\columnwidth]{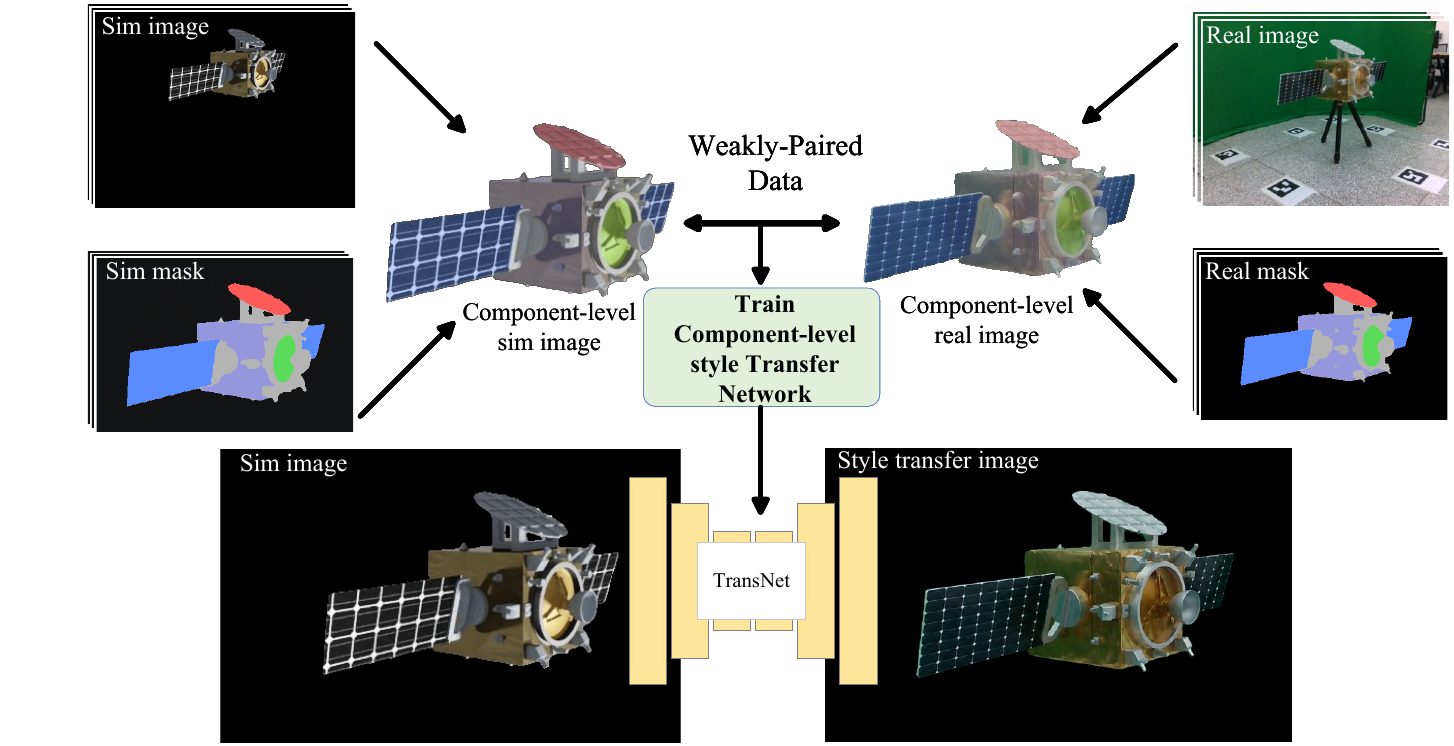}
    \caption{
    Illustration of the proposed component-level Sim2Real style transfer framework.
    Weakly paired real--synthetic samples and component masks are used to train a component-level style transfer network, which translates synthetic satellite images into realistic images while preserving structural annotations.
    }
    \label{fig:3}
\end{figure}

Several Sim2Real strategies attempt to reduce this gap. Domain randomization improves robustness by varying rendering parameters such as texture, lighting, viewpoint, and background~\cite{tobin2017domainRandomization,tremblay2018domainRandomization}; photorealistic rendering instead tries to make the simulator closer to the physical imaging process. Image-level translation provides another route: paired and unpaired methods such as pix2pix, CycleGAN, DRIT, MUNIT, and CUT translate synthetic images toward the real domain with different assumptions about correspondence and style diversity~\cite{isola2017pix2pix,cyclegan,lee2018diverse,huang2018munit,park2020contrastive}. Region-adaptive synthesis methods such as SPADE and SEAN further show that semantic layouts can control where style information is injected~\cite{park2019spade,zhu2020sean}. These methods are useful, but a satellite training image has a stricter requirement than visual plausibility. Its translated appearance must remain aligned with the pose, mask, and component annotations inherited from simulation.

This requirement exposes a limitation of treating satellite Sim2Real as generic full-image translation. A satellite is a composition of visually heterogeneous parts. Solar panels, metallic bus surfaces, antennas, docking structures, and nozzles do not share the same radiometric behavior, and a single global style code can easily mix their statistics. In addition, full-image translation may alter the background, blur component boundaries, or introduce texture leakage across parts. Such changes are not merely cosmetic: they weaken the correspondence between the translated image and the original synthetic supervision. For downstream pose estimation, preserving the object silhouette, component layout, high-frequency boundaries, and pose consistency is as important as reducing the distributional distance to real images.

To address this problem, we propose a component-level structure-preserving style transfer framework for satellite Sim2Real dataset construction. As shown in Fig.~\ref{fig:3}, the framework first constructs weakly paired real--synthetic samples through real image acquisition, ArUco-based camera-pose measurement~\cite{garrido2014aruco}, component mask construction, and CAD-based rendering. It then extracts part-wise real-domain style representations from real satellite images and injects them into the corresponding synthetic component regions through mask-aligned modulation. Structure-preserving losses are used to constrain the translation so that the generated images become more realistic while retaining the geometry and annotations of the synthetic source. The resulting dataset is evaluated by training a downstream GDRNet pose estimator~\cite{wang2021gdrnet} on translated synthetic images and testing it on real satellite observations from the same calibrated target domain.

The main contributions of this work are:
\begin{itemize}
    \item A weakly paired real--synthetic construction pipeline for satellite Sim2Real style-transfer training, combining calibrated real acquisition, component mask construction, CAD-based rendering, and viewpoint-consistent synthetic rendering.
    \item A component-level, mask-aligned style transfer method that models part-specific real-domain appearance and translates synthetic satellite images while preserving simulation-derived geometric annotations.
    \item A downstream pose-estimation evaluation in the constructed calibrated target domain, showing that the proposed dataset synthesis method improves real-image performance over representative image-translation baselines.
\end{itemize}

\section{Related Work}
\label{sec:related-work}

\subsection{Satellite Pose Estimation and Synthetic Training Data}

Image-based spacecraft pose estimation has been studied for non-cooperative rendezvous and close-proximity operations, where active markers or cooperative communication may be unavailable~\cite{opromolla2017review,song2022relativeNavigationSurvey}. Early learning-based systems such as Spacecraft Pose Network demonstrated that neural networks can estimate spacecraft pose from images, and the SPEED dataset and satellite pose-estimation challenges established common benchmarks for this task~\cite{sharma2020spn,kisantal2020speedChallenge}. Landmark-regression and refinement pipelines, as well as photorealistic rendering for pose training, further show how geometric cues and synthetic imagery are combined in this field~\cite{chen2019landmarkSatellitePose,proenca2020photorealistic}. SPEED+ later made the domain gap more explicit by adding hardware-in-the-loop images with substantially different appearance from synthetic training data~\cite{park2022speedPlus}. Recent surveys and challenge analyses report that deep spacecraft pose estimators still depend heavily on synthetic imagery, domain adaptation, and online refinement when real labeled data are scarce~\cite{pauly2023surveySpacecraftPose,park2024spnv2}.

Most spacecraft pose-estimation work focuses on the estimator, the benchmark, or the adaptation strategy applied to the task network. In contrast, this paper focuses on the training data before the pose network is trained. We use a pose estimator as the downstream evaluator; the main contribution is a component-aware synthetic-to-real image generation process that keeps simulation-derived annotations usable for training.

\subsection{Synthetic Data and Sim2Real Adaptation}

Synthetic data are widely used in pose and viewpoint estimation because rendering from 3D models can provide dense labels at scale. Rendered views have been used to train CNNs for viewpoint estimation~\cite{su2015renderForCNN}, and spacecraft-specific work has developed synthetic data generation and validation pipelines for image-based pose estimation~\cite{bechini2023spacecraftDataset}. The appeal is clear: a CAD model can yield accurate object pose, foreground mask, bounding box, and component-level labels without manual annotation. The drawback is equally clear in satellite settings, where illumination, material response, camera noise, and background statistics differ from the physical acquisition environment~\cite{park2022speedPlus,pauly2023surveySpacecraftPose}.

Sim2Real adaptation addresses this discrepancy in several ways. Domain randomization deliberately varies rendering parameters so that the trained model becomes less sensitive to any single simulated appearance~\cite{tobin2017domainRandomization,tremblay2018domainRandomization}; pose systems trained with synthetic data have used this principle to improve transfer in robotic object pose estimation~\cite{tremblay2018dope}. Render-and-compare methods such as MegaPose also illustrate the value of large-scale rendered hypotheses when real pose annotations are limited~\cite{labbe2023megapose}. More broadly, recent 3D perception and robotic manipulation studies have used point/voxel feature learning and generative models to improve geometric understanding and action generation~\cite{xie2024psvmlp,xie2025dexmgnet}. Other approaches improve rendering realism or adapt the task network through multi-task learning, online refinement, or semantic decomposition~\cite{li2021sdpose,park2024spnv2}. These strategies can reduce the synthetic-to-real gap, but they do not directly solve the problem considered here: when synthetic images are translated into a real-like style, the translated image must still match the original pose, mask, and component annotations. Our work therefore treats synthetic data not only as a source of visual samples, but as a source of geometric supervision that must be protected during translation.

\subsection{Image Translation for Sim2Real Dataset Construction}

Image-to-image translation methods provide a natural tool for making synthetic images look more realistic. pix2pix established a paired conditional-GAN formulation, CycleGAN introduced cycle-consistent unpaired translation, DRIT and MUNIT disentangled content and style for diverse outputs, and CUT used contrastive learning to preserve local correspondence during unpaired translation~\cite{isola2017pix2pix,cyclegan,lee2018diverse,huang2018munit,park2020contrastive}. For pose-estimation data construction, instance-level Sim2Real style transfer has also been used to improve the realism of training images while retaining object-level supervision~\cite{ikeda2022sim2real}. These methods motivate our use of image-level adaptation, but they usually operate at the full-image or object-instance level.

For satellite data construction, full-image translation is often too coarse. It may change the background, shift illumination globally, or mix the appearance statistics of solar panels, metallic bodies, antennas, and nozzles. Even when the translated image is visually plausible, these changes can weaken the consistency between the image and its inherited synthetic annotations. Our method keeps the translation on the satellite foreground and separates appearance transfer by component, so the generated image is designed for downstream annotation-preserving training rather than unconstrained visual realism.

\subsection{Component-Aware and Structure-Preserving Translation}

Semantic image synthesis shows that spatial labels can guide image generation more precisely than a global latent code. SPADE uses semantic layouts to modulate normalization layers, and SEAN extends this idea with region-adaptive style control~\cite{park2019spade,zhu2020sean}. These methods provide useful building blocks for region-aware generation, but they are not designed specifically for satellite Sim2Real data construction. In particular, they do not by themselves enforce that a translated CAD-rendered image remains aligned with the original pose and component annotations.

Structure preservation is therefore a separate requirement. CUT's PatchNCE objective encourages local feature correspondence during translation~\cite{park2020contrastive}, and ADD-based pose metrics ultimately evaluate whether geometric prediction remains accurate in the target domain~\cite{hinterstoisser2012linemod}. In our framework, component masks define where each real-domain style code is injected, while PatchNCE, self-regularization, and edge-consistency losses constrain how much the synthetic structure can change. The resulting formulation combines region-aware style modulation with annotation-preserving constraints, which is the main distinction from generic semantic image synthesis or full-image Sim2Real translation.

\begin{figure*}[!t] 
    \centering 
    \includegraphics[width=\textwidth]{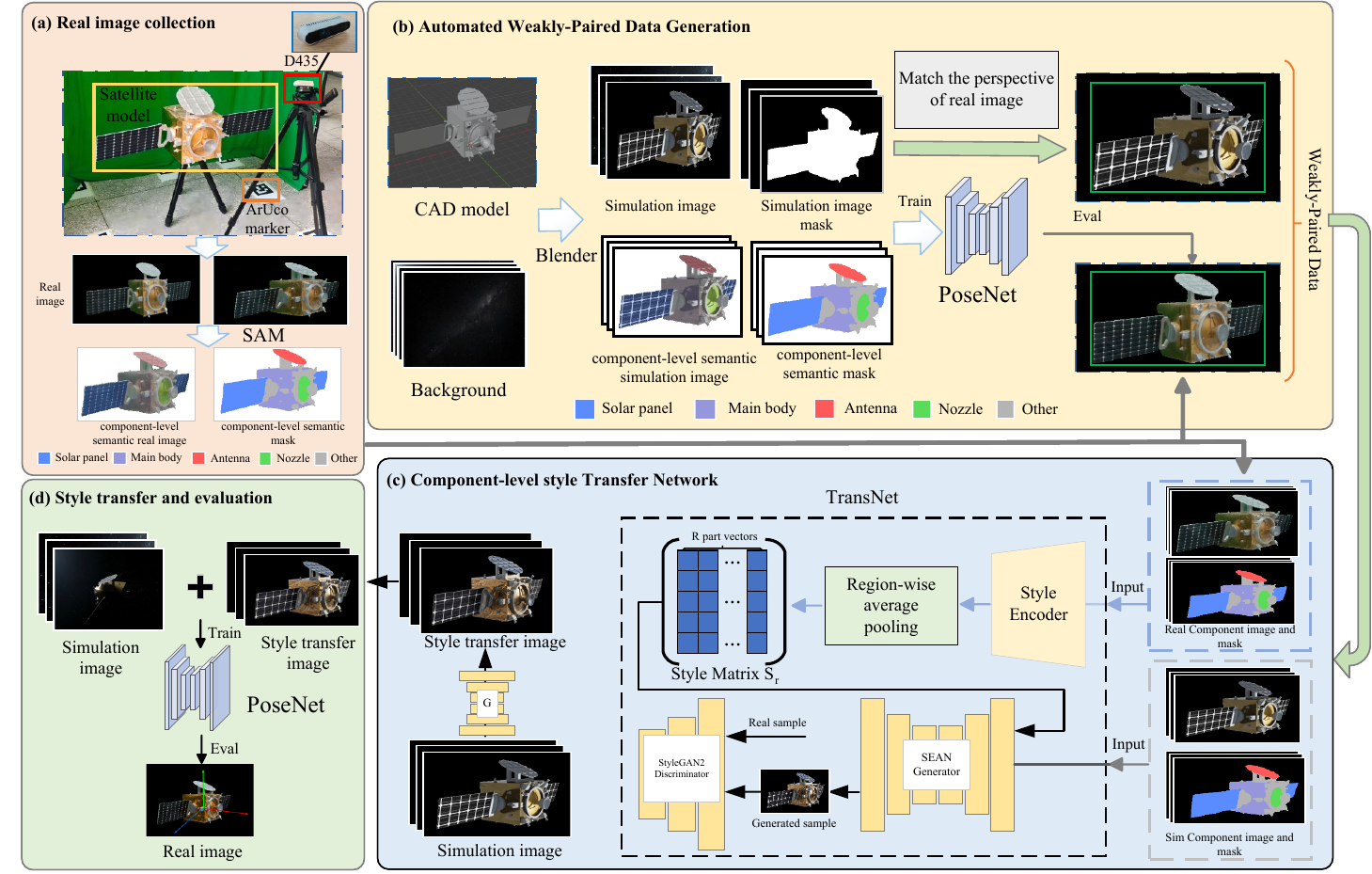}
    \caption{
    Overview of the proposed framework.
    (a) Real satellite images and component masks are acquired in a calibrated environment.
    (b) Weakly paired real--synthetic samples are generated using ArUco-based camera-pose measurement, CAD-based rendering, and view consistency filtering.
    (c) TransNet, the proposed component-level style transfer network, extracts part-wise real style codes and synthesizes structure-preserving translated images.
    (d) The translated images are used for realistic dataset synthesis and downstream pose-estimation evaluation.
    }
    \label{fig:1} 
\end{figure*}

Our method differs from these approaches by introducing component-level semantic constraints and structure-preserving objectives. It learns part-wise real-domain style representations and injects them into corresponding synthetic regions, thereby improving realism while preserving satellite geometry, component layout, and annotation consistency.

\section{Method}
\label{sec:methodology}

We propose a part-aware structure-preserving style transfer framework for satellite Sim2Real dataset construction. Given large-scale synthetic images and a limited number of real images, our goal is to synthesize realistic satellite images that better match real observations while preserving the geometric annotations inherited from simulation. The key assumption is that the style transfer network requires substantially fewer real samples than downstream vision models. To this end, we introduce component-level semantic constraints into the translation process, enabling fine-grained appearance modeling for different satellite parts while preventing structural degradation during style transfer. For brevity, the proposed component-level style transfer network is denoted as TransNet in the figures and the following descriptions.

As shown in Fig.~\ref{fig:1}, the framework consists of four stages. First, real satellite images are captured in a calibrated environment, where ArUco markers provide camera-pose references and camera-frame satellite poses for evaluation, and component-level masks are produced with SAM-assisted segmentation~\cite{kirillov2023sam}. Second, synthetic images and component masks are rendered from the satellite CAD model in Blender using the measured camera viewpoints, followed by view consistency filtering to retain weakly paired real--synthetic samples. Third, the component-level style transfer network extracts part-wise style codes from real images and injects them into a SEAN-based generator to translate synthetic images while preserving geometry. Finally, the trained network is applied to large-scale synthetic data, and the generated dataset is evaluated through downstream satellite pose estimation. PoseNet is used as the synthetic-trained Non-Adaptation baseline in the downstream evaluation rather than as the source of ground-truth poses for weak-pair construction.

\subsection{Real Image Acquisition and Reference Data Construction}

As illustrated in Fig.~\ref{fig:1}(a), real reference data are collected in a calibrated environment using an Intel RealSense D435 camera, a physical satellite model, and multiple ArUco markers~\cite{garrido2014aruco}. The markers are calibrated in a common environment coordinate system, which allows the camera pose of each captured image to be recorded. Since the satellite model is also registered to this coordinate frame, the marker-based calibration provides the satellite pose in the camera coordinate system. This camera-frame satellite pose is used as the ground-truth reference for downstream pose evaluation.

In addition to pose references, component-level semantic masks are constructed for real images. We use SAM-assisted segmentation~\cite{kirillov2023sam} to obtain masks for satellite components such as the main body, solar panels, antennas, and nozzles. These masks provide real-domain component guidance for part-aware style representation learning. To improve viewpoint coverage, real images are captured from multiple viewing angles.
\subsection{Automated Weakly-Paired Data Generation}

As shown in Fig.~\ref{fig:1}(b), we automatically construct weakly paired real--synthetic samples for style transfer training without pixel-level registration. Given the ArUco-recovered camera pose in the world coordinate system and the satellite CAD model, Blender renders a synthetic view with a similar camera viewpoint and object layout.

\begin{figure*}[!t] 
    \centering 
    \includegraphics[width=\textwidth]{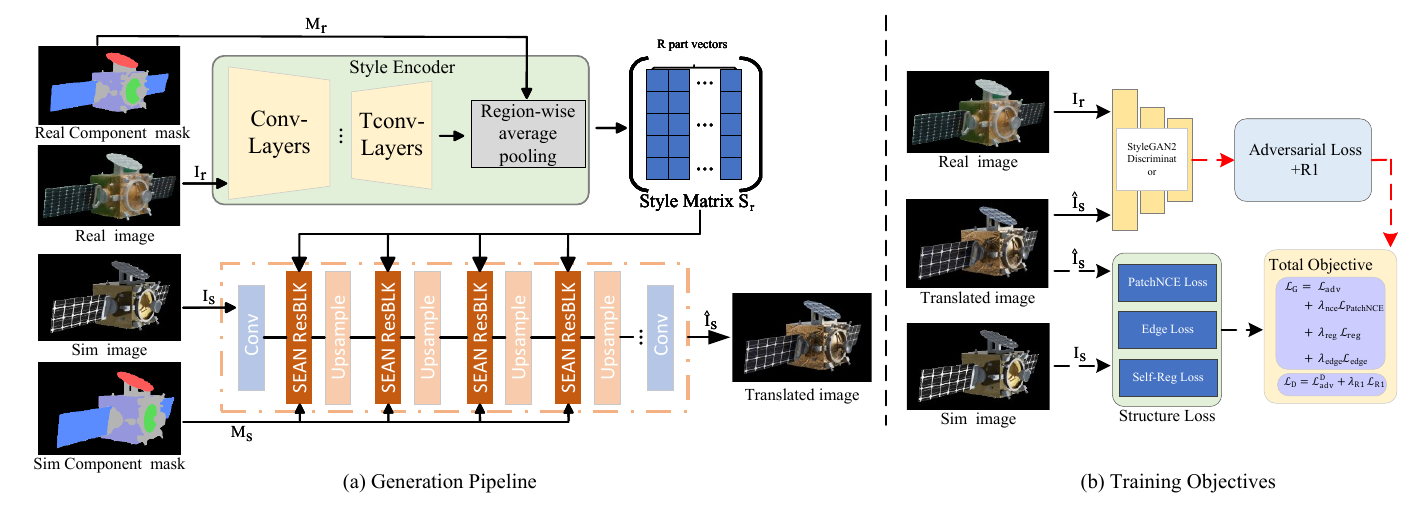}
    \caption{
    Detailed architecture of the proposed component-level structure-preserving style transfer network.
    (a) Generation pipeline, where real component styles are encoded into a part-wise style matrix and injected into the synthetic image through a SEAN-based generator.
    (b) Training objectives, including adversarial supervision and structure-preserving losses.
    }
    \label{fig:2} 
\end{figure*}

We first render a synthetic training set from the CAD model, where object poses and component-level masks are automatically generated. An initial PoseNet is trained solely on this synthetic set and is retained as the Non-Adaptation baseline in the downstream pose-estimation experiment. For weak-pair construction, the Blender virtual camera is driven by the measured camera pose rather than by style-transfer outputs, so the rendered synthetic image keeps the viewpoint coverage of the real acquisition sequence.

To reduce mismatched pairs caused by unreliable pose predictions, we apply a view-wise instance consistency filter. For each real view \(i\), we record the number of detected satellite instances as \(n_i\). Since our acquisition setup contains a single target, valid samples should satisfy \(n_i=\bar{n}\), where \(\bar{n}=1\). We define
\[
m_i =
\begin{cases}
1, & n_i = \bar{n}, \\
0, & n_i \neq \bar{n},
\end{cases}
\qquad \bar{n}=1 ,
\]
where \(m_i\) denotes whether the \(i\)-th view is retained. Samples with \(m_i=0\) are discarded as mismatched samples.

The retained real--synthetic pairs share similar viewpoints, object layouts, and coarse pose configurations, but may still contain residual alignment errors. We therefore define them as weakly paired samples rather than strictly aligned pairs. In the current dataset, 60 weak pairs are retained to cover the observed viewing range and to provide geometrically consistent training data for the subsequent appearance translation stage.

\subsection{Component-Level Structure-Preserving Style Transfer}

For satellite Sim2Real dataset construction, style transfer should improve the visual realism of synthetic images while preserving the annotation consistency inherited from simulation. Since the translated images are used to train downstream vision models, their object masks, component layouts, bounding boxes, and poses should remain aligned with the original synthetic annotations. Therefore, our goal is to synthesize realistic satellite images without changing the object silhouette, component layout, or structural boundaries.

Satellite targets contain multiple components with distinct material properties, such as solar panels, metallic main bodies, antennas, docking rings, and nozzles. A global instance-level style code may entangle the appearance statistics of different components, leading to cross-component texture contamination. To address this problem, we model each satellite component as an independent appearance unit and perform mask-aligned component-wise style transfer.

\begin{figure*}[!t]
    \centering
    \includegraphics[width=\textwidth]{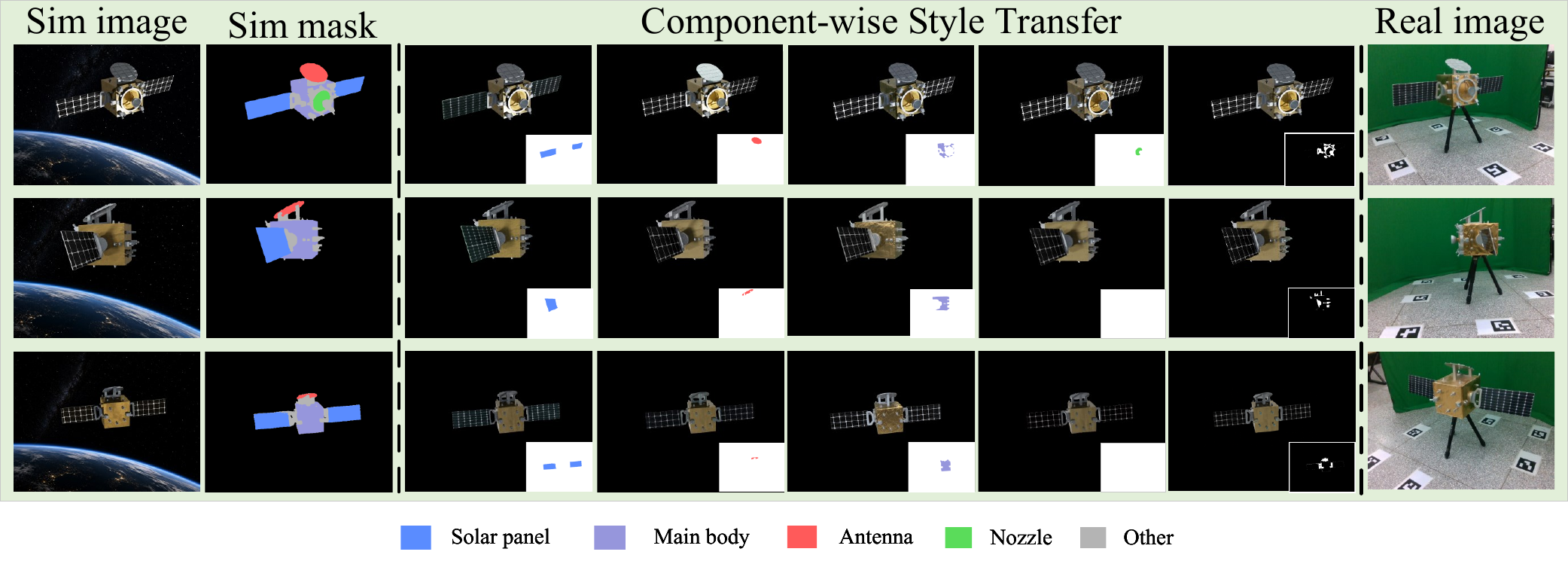}
    \caption{
    Component-wise style transfer visualization.
    Each intermediate result transfers the style of one selected satellite component while keeping the remaining components unchanged.
    }
    \label{fig:component_control}
\end{figure*}

As shown in Fig.~\ref{fig:1}(c) and Fig.~\ref{fig:2}(a), given a real reference image \(I_r\) and its component mask \(M_r\), TransNet first uses a style encoder \(E_s\) to extract real-domain appearance features. The component mask partitions the satellite into \(R\) semantic regions, and region-wise masked pooling is used to obtain a part-wise style matrix:
\[
S_r = E_s(I_r, M_r) \in \mathbb{R}^{C \times R},
\]
where \(C\) denotes the style feature dimension and each column of \(S_r\) represents the style code of one satellite component. In our implementation, \(C=512\). The real component mask \(M_r\) defines how real-domain appearance statistics are decomposed, while the synthetic component mask \(M_s\) determines where the style codes are injected.

Given a synthetic image \(I_s\), its component mask \(M_s\), and the real style matrix \(S_r\), the generator synthesizes a translated image:
\[
\hat{I}_s = G(I_s, M_s, S_r).
\]

We adopt a SEAN-based generator as the translation backbone. At each modulation layer \(l\), the synthetic component mask is resized to the corresponding feature resolution, denoted as \(M_s^l\), and used to route each component style code to its matching synthetic region. The modulation parameters are predicted as:
\[
\gamma_l, \beta_l = \Phi_l(M_s^l, S_r),
\]
and the intermediate feature map \(F_l\) is modulated by
\[
\hat{F}_l = \gamma_l \odot \mathrm{Norm}(F_l) + \beta_l ,
\]
where \(\odot\) denotes element-wise multiplication. This mask-aligned modulation injects each real component style only into the corresponding synthetic component region, reducing texture leakage while preserving the synthetic pose, silhouette, and part layout.

Fig.~\ref{fig:component_control} visualizes the component-wise controllability of the proposed style representation. By selectively transferring the style of one semantic component, the network changes the appearance of the corresponding region while keeping the remaining components largely unchanged. This demonstrates that the learned representation supports fine-grained part-level appearance control.

As shown in Fig.~\ref{fig:2}(b), the network is trained with adversarial supervision and structure-preserving constraints. A StyleGAN2 discriminator \(D\) is used to encourage translated images to match the real image distribution, with R1 regularization for stable training~\cite{karras2020stylegan2}:
\[
\mathcal{L}_{D}
=
\mathcal{L}_{\mathrm{adv}}^{D}
+
\lambda_{\mathrm{R1}}\mathcal{L}_{\mathrm{R1}} .
\]

To preserve the validity of synthetic annotations, we further impose structure-preserving losses between the synthetic input \(I_s\) and the translated output \(\hat{I}_s\). PatchNCE encourages local feature correspondence, self-regularization penalizes excessive pixel-level changes, and edge loss preserves high-frequency boundary cues. The generator objective is:
\[
\mathcal{L}_{G}
=
\mathcal{L}_{\mathrm{adv}}
+
\lambda_{\mathrm{nce}}\mathcal{L}_{\mathrm{PatchNCE}}
+
\lambda_{\mathrm{reg}}\mathcal{L}_{\mathrm{reg}}
+
\lambda_{\mathrm{edge}}\mathcal{L}_{\mathrm{edge}} .
\]

The PatchNCE loss follows the local contrastive form used in contrastive unpaired translation. For feature layer \(l\in\mathcal{L}\) and sampled patch index \(p\), let \(q_l^p\) denote the projected feature of the translated image, \(k_l^{p,+}\) the corresponding positive feature from the synthetic input, and \(k_l^{p',-}\) the features of other sampled patches. With temperature \(\tau\), the loss is
\[
\begin{aligned}
\mathcal{L}_{\mathrm{PatchNCE}}
&=
\frac{1}{|\mathcal{L}|}
\sum_{l\in\mathcal{L}}
\frac{1}{P_l}
\sum_{p=1}^{P_l}
\ell_l^p,\\
\ell_l^p
&=
-\log
\frac{\exp(q_l^p \cdot k_l^{p,+}/\tau)}
{Z_l^p},\\
Z_l^p
&=
\exp(q_l^p \cdot k_l^{p,+}/\tau)
+
\sum_{p'\neq p}\exp(q_l^p \cdot k_l^{p',-}/\tau) .
\end{aligned}
\]
This term discourages local geometric drift by keeping corresponding synthetic and translated patches close in the feature space.

The self-regularization loss is applied on the satellite foreground:
\[
\mathcal{L}_{\mathrm{reg}}
=
\frac{
\left\|M_s^{obj}\odot(\hat{I}_s-I_s)\right\|_1
}{
\left\|M_s^{obj}\right\|_1+\epsilon
},
\]
where \(M_s^{obj}\) is the foreground mask and \(\epsilon\) avoids division by zero. For edge consistency, we compute Sobel edge maps \(E(\cdot)\) and penalize foreground boundary changes:
\[
\mathcal{L}_{\mathrm{edge}}
=
\frac{
\left\|M_s^{obj}\odot(E(\hat{I}_s)-E(I_s))\right\|_1
}{
\left\|M_s^{obj}\right\|_1+\epsilon
}.
\]

In our implementation, synthetic component masks are rendered from the CAD model, while real component masks are obtained using SAM-based automatic segmentation with refinement. This design enables scalable generation of realistic satellite training data while preserving the structural and annotation consistency required by downstream Sim2Real tasks.

\subsection{Style Transfer and Evaluation}

\begin{figure}[h]
    \centering
    \includegraphics[width=\linewidth]{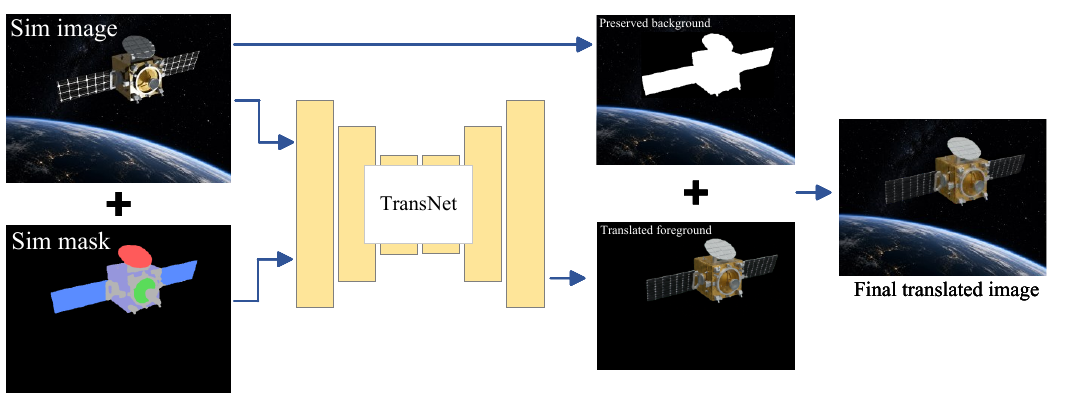}
    \caption{
    Foreground-only style transfer during inference.
    TransNet translates the satellite foreground conditioned on the synthetic image and component mask, while the original background is preserved through mask-based composition.
    }
    \label{fig:7}
\end{figure}

After training, the component-level style transfer network is applied to large-scale synthetic satellite images for realistic dataset synthesis. As shown in Fig.~\ref{fig:7}, given a synthetic image \(I_s\), its component mask \(M_s\), and a sampled real-domain style matrix \(S_r\), the generator produces a translated foreground image:
\[
\tilde{I}_s = G(I_s, M_s, S_r),
\]
where \(S_r\) is sampled from the learned component-wise style set extracted from real satellite observations.

To preserve the original scene layout and avoid background artifacts, only the satellite foreground is translated. The final image is obtained by mask-based composition:
\[
\hat{I}_s
=
M_s^{obj} \odot \tilde{I}_s
+
(1 - M_s^{obj}) \odot I_s ,
\]
where \(M_s^{obj}\) is the binary object mask derived from \(M_s\), and \(\odot\) denotes element-wise multiplication. In this way, the satellite appearance is transferred toward the real domain while the background remains unchanged.

The generated images inherit the annotations of the original synthetic data, including object masks, component masks, bounding boxes, and object poses. Since the translation is conditioned on the synthetic structure and constrained by PatchNCE, self-regularization, and edge-consistency losses during training, the translated images preserve the object silhouette, component layout, and structural boundaries.

For evaluation, we use satellite pose estimation as the downstream Sim2Real task. A GDRNet pose estimator is trained on either original synthetic images or translated synthetic images and evaluated on the same real test set. Improved performance on real images indicates that the generated data reduce the visual domain gap while preserving geometric supervision from simulation.

\section{Experiments}

In this section, we evaluate the proposed component-level structure-preserving style transfer framework for satellite Sim2Real dataset construction. We first describe the datasets, implementation settings, baselines, and metrics. We then compare our method with representative translation baselines in terms of image fidelity, qualitative results, and downstream pose-estimation performance. Finally, ablation studies are conducted to analyze the effects of component mask guidance and structure-preserving losses.

\subsection{Experimental Setup}

We evaluate our method on a synthetic satellite dataset rendered from a self-built CAD model and a real satellite dataset captured in a calibrated environment. As shown in Fig.~\ref{fig:1}(a), the real acquisition setup consists of an Intel RealSense D435 camera, a physical satellite model, and eight ArUco markers for camera calibration and pose reference construction. We collect 100 real images from multiple viewpoints at a distance of approximately 3--5 m. These real images are used as unlabeled target-domain references for style transfer training and as test images for final evaluation, but their pose labels are not used to optimize the downstream pose estimator. This protocol should be interpreted as target-domain Sim2Real data construction rather than a held-out unseen-domain generalization test.

\begin{figure}[h]
    \centering
    \includegraphics[width=\linewidth]{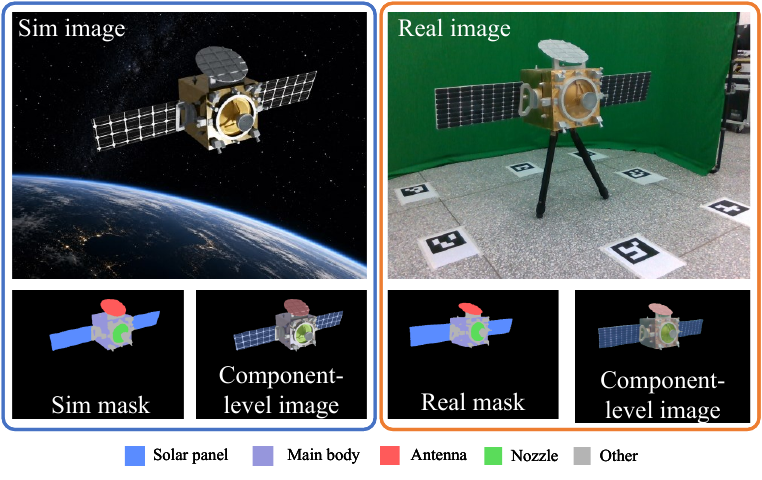}
    \caption{
    Data samples and component masks. Synthetic and real images share the same component taxonomy, including solar panels, main body, antenna, nozzle, and other minor structures.
    }
    \label{fig:5}
\end{figure}

The synthetic dataset contains 5,000 Blender-rendered images at a resolution of \(640 \times 480\), with a 7:3 training-validation split. The virtual camera distance is also set to approximately 3--5 m to match the real acquisition setting. Space-image backgrounds and a single-side point light source are used for rendering. Each synthetic image is automatically annotated with object pose, bounding box, foreground mask, and component-level semantic masks.

Fig.~\ref{fig:5} shows representative synthetic and real samples with their component masks. Both domains share the same component taxonomy, including solar panels, main body, antenna, nozzle, and other minor structures. Synthetic component masks are rendered from the CAD model, while real component masks are obtained using SAM-assisted automatic segmentation~\cite{kirillov2023sam}. This consistent component definition enables component-aware style transfer across the synthetic and real domains.

All experiments are conducted on a workstation equipped with a single NVIDIA RTX 4090 GPU. The proposed style-transfer network is trained for 100 epochs with batch size 8, Adam optimizer, learning rate \(2\times10^{-4}\), and \(512\times512\) input resolution. The loss weights are set to \(\lambda_{\mathrm{R1}}=10.0\), \(\lambda_{\mathrm{nce}}=1.0\), \(\lambda_{\mathrm{reg}}=0.1\), and \(\lambda_{\mathrm{edge}}=0.5\). PatchNCE uses layers 1--4 as the multi-scale feature set, with temperature \(\tau=0.07\).

For downstream Sim2Real evaluation, we use GDRNet as the pose estimator and evaluate all models on the same real test set. GDRNet is trained for 150 epochs with batch size 32, Adam optimizer, learning rate \(1\times10^{-4}\), and \(640\times480\) input resolution. The Non-Adaptation setting uses the original synthetic training split, while each adaptation method uses the translated version of the same synthetic training split, approximately 3,500 images. All downstream pose-estimation runs use a fixed random seed of 42 and are reported as a single deterministic run.

\begin{figure*}[t]
    \centering
    \includegraphics[width=\textwidth]{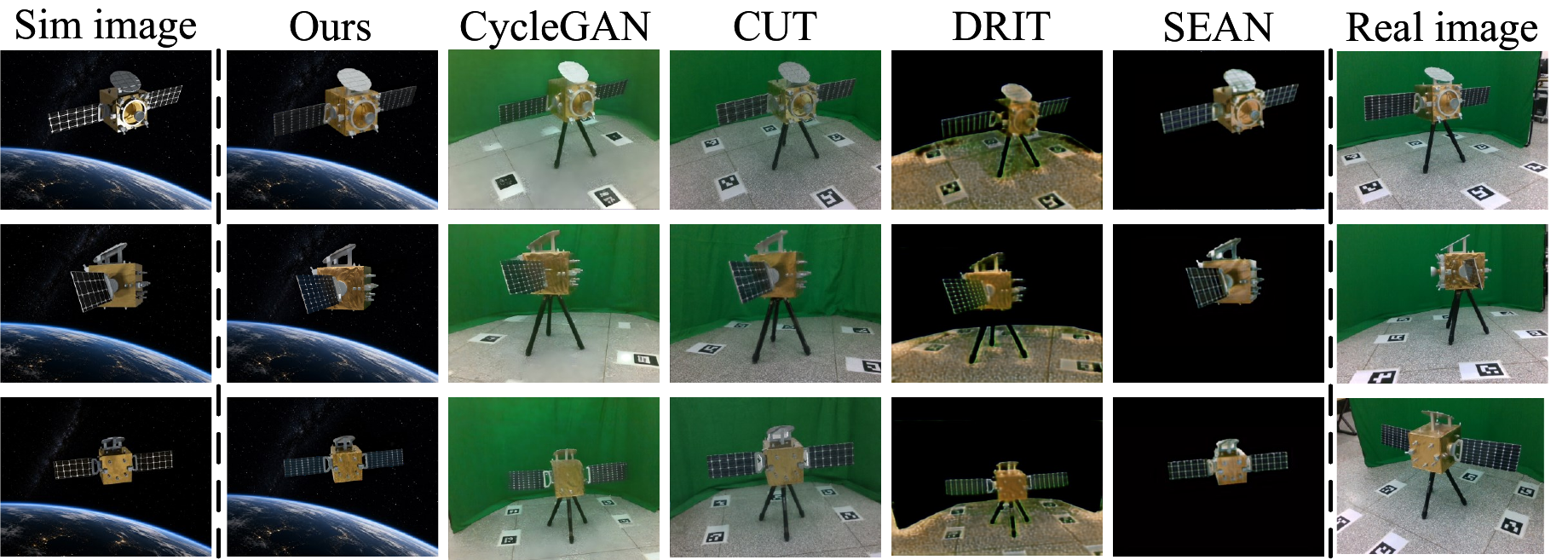}
    \caption{
    Qualitative comparison of different Sim2Real translation methods.
    CycleGAN, CUT, and DRIT perform full-image translation and may modify both the satellite and background.
    SEAN uses semantic masks for region-adaptive translation.
    Our method better preserves the original background, satellite geometry, and component layout while improving real-domain appearance.
    }
    \label{fig:9}
\end{figure*}

The compared methods include Non-Adaptation, CycleGAN~\cite{cyclegan}, DRIT~\cite{lee2018diverse}, CUT~\cite{park2020contrastive}, SEAN~\cite{zhu2020sean}, and our method. Because these translation models have different architectures and training objectives, the comparison is controlled at the data and evaluation levels rather than by forcing identical optimizer schedules. All methods use the same synthetic training split and the same real reference set where applicable. Their translated outputs are resized to the same resolution and then used to train the same GDRNet model under the same downstream protocol. For methods that perform full-image translation, the output image is used directly; for the proposed method, foreground-only translation is composed with the original synthetic background as described in Sec.~\ref{sec:methodology}. This setting evaluates whether each data-construction strategy produces synthetic-derived images that are useful for real-domain pose estimation.

\subsection{Evaluation Metrics}

For image-level evaluation, we report FID~\cite{heusel2017fid} and KID~\cite{binkowski2018kid} between translated synthetic images and real observations. In our evaluation, the translated synthetic validation set, approximately 1,000 images, is compared with the 100 real images. Both metrics are computed from pretrained Inception-v3 features; FID measures the Frechet distance between Gaussian feature statistics, while KID estimates a kernel distance between feature distributions.

For downstream pose evaluation, we use ADD over the satellite model point set \(\mathcal{M}\):
\[
\mathrm{ADD}
=
\frac{1}{|\mathcal{M}|}
\sum_{x\in\mathcal{M}}
\left\|
(Rx+t)-(\hat{R}x+\hat{t})
\right\|_2 ,
\]

where \(R,t\) are the reference pose and \(\hat{R},\hat{t}\) are the predicted pose. The ADD pass rate is reported at the \(0.02\,\mathrm{m}\) threshold, meaning that a prediction is counted as correct when its average model-point distance is below 2 cm. AUC is computed from the ADD pass-rate curve over thresholds from \(0\) to \(0.10\,\mathrm{m}\).

For structural consistency in the ablation study, Mask IoU compares the component mask inferred from the translated result with the original synthetic component mask, rather than with a real-image mask. Edge Error is computed as the foreground-masked Sobel edge difference between the translated image and the synthetic source, normalized by the number of foreground pixels. These two metrics evaluate whether the translation preserves the component layout and high-frequency boundaries needed for annotation-preserving training.

\subsection{Comparative Experiments}

We compare the proposed method with representative baselines from two perspectives: image-level style transfer quality and downstream Sim2Real performance.

\textbf{Qualitative comparison.}
We first compare the visual quality and structural consistency of different Sim2Real translation methods. As shown in Fig.~\ref{fig:9}, CycleGAN, CUT, and DRIT perform full-image translation and therefore modify both the satellite foreground and the background. Although these methods introduce real-domain appearance cues, they also change the original space background and may weaken the consistency between the translated images and the synthetic annotations. In particular, the translated results show global color and illumination changes, and the satellite components are not explicitly controlled by their semantic regions.

SEAN uses semantic masks for region-adaptive modulation and thus better preserves the foreground-background separation than full-image translation methods. However, its results still show limited component-specific realism and less stable fine details around solar panels, antennas, and the main body. In contrast, our method preserves the original synthetic background and satellite geometry while transferring realistic appearance to the corresponding components. The solar panels, main body, and small structures remain better aligned with the input layout, indicating that the proposed component-aware and structure-preserving translation is more suitable for annotation-preserving Sim2Real dataset synthesis.

\textbf{Image fidelity evaluation.}
We further evaluate image-level realism using FID and KID between translated satellite images and real observations. As shown in Tab.~\ref{tab:image_fidelity}, our method achieves the lowest FID and KID, with scores of 54.32 and 0.048, respectively. Compared with the strongest baseline CUT, our method reduces FID from 62.85 to 54.32 and KID from 0.056 to 0.048, indicating that the proposed component-aware translation better matches the real image distribution. In contrast, DRIT produces the largest distribution discrepancy, suggesting that image-level translation without
explicit structure-preserving constraints may be insufficient for satellite Sim2Real dataset construction.

FID and KID are used as auxiliary indicators of image-level realism, since they measure distributional similarity but do not directly verify whether geometric annotations remain valid after translation. Therefore, downstream pose-estimation performance is used as the task-level criterion for evaluating Sim2Real effectiveness.

\begin{table}[h]
\centering
\caption{
FID and KID between translated satellite images and real observations.
}
\label{tab:image_fidelity}
\begin{tabular}{lcc}
\hline
Method & FID \(\downarrow\) & KID \(\downarrow\) \\
\hline
CycleGAN~\cite{cyclegan} & 78.64 & 0.071 \\
DRIT~\cite{lee2018diverse} & 117.41 & 0.111 \\
CUT~\cite{park2020contrastive} & 62.85 & 0.056 \\
SEAN~\cite{zhu2020sean} & 93.29 & 0.085 \\
Ours & \textbf{54.32} & \textbf{0.048} \\
\hline
\end{tabular}
\end{table} 

\textbf{Downstream pose evaluation.}
To evaluate the Sim2Real effectiveness of the generated data, we train the same GDRNet pose estimator using different training sets and evaluate all models on the same real test set. The compared settings include Non-Adaptation, CycleGAN~\cite{cyclegan}, DRIT~\cite{lee2018diverse}, CUT~\cite{park2020contrastive}, SEAN~\cite{zhu2020sean}, and our method. For a fair comparison, all translation methods use the same synthetic training split and the same real reference set, while paired information is used only by methods that can explicitly exploit it.

\begin{figure}[h]
    \centering
    \includegraphics[width=\linewidth]{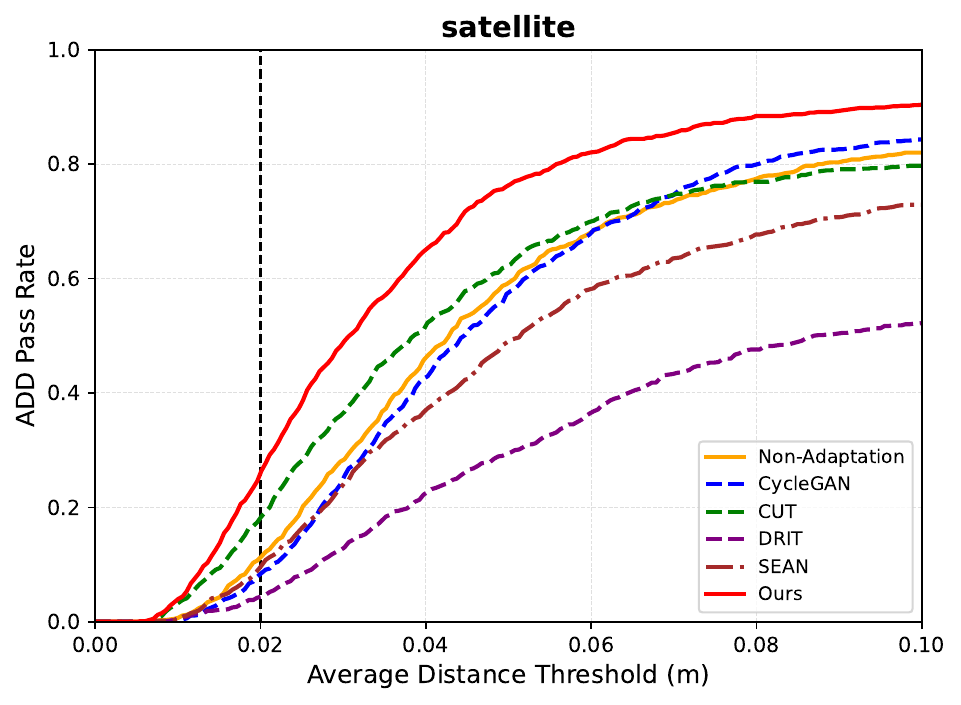}
    \caption{
    Average distance threshold curves of Non-Adaptation, our method, and other adaptation methods.
    The dashed vertical line denotes the \(0.02\,\mathrm{m}\) threshold used for the ADD values in Tab.~\ref{tab:pose_results}.
    }
    \label{fig:4}
\end{figure}

Fig.~\ref{fig:4} shows the average distance threshold curves, and Tab.~\ref{tab:pose_results} reports the ADD pass rate at the \(0.02\,\mathrm{m}\) threshold and the AUC score. Our method achieves the highest ADD pass rate of 0.260 and the highest AUC of 0.611. Compared with the strongest baseline CUT, our method improves the ADD pass rate from 0.182 to 0.260 and the AUC from 0.513 to 0.611. These results demonstrate that the proposed component-aware and structure-preserving translation improves downstream pose estimation by reducing the synthetic-to-real appearance gap while preserving useful geometric cues.

\begin{table}[h]
\centering
\caption{
ADD pass rate at \(0.02\,\mathrm{m}\) and area under the curve (AUC) for downstream pose estimation.
}
\label{tab:pose_results}
\begin{tabular}{lcc}
\hline
Training Data & ADD \(\uparrow\) & AUC \(\uparrow\) \\
\hline
Non-Adaptation & 0.111 & 0.486 \\
CycleGAN~\cite{cyclegan} & 0.085 & 0.481 \\
DRIT~\cite{lee2018diverse} & 0.044 & 0.274 \\
CUT~\cite{park2020contrastive} & 0.182 & 0.513 \\
SEAN~\cite{zhu2020sean} & 0.098 & 0.417 \\
Ours & \textbf{0.260} & \textbf{0.611} \\
\hline
\end{tabular}
\end{table}

\subsection{Ablation Study}

We conduct ablation studies on four removable components in the implemented framework: component mask guidance, PatchNCE loss, pixel-level self-regularization, and edge-consistency loss. Since some ablated variants may weaken the alignment between translated images and the original synthetic annotations, we mainly evaluate the ablation results using visual comparison and image-level structural consistency, rather than training a downstream pose estimator for every variant. The downstream pose experiment in Tab.~\ref{tab:pose_results} is therefore used to validate the full data-construction pipeline, while the ablation study isolates which translation constraints most affect image fidelity and structural preservation.

\begin{figure*}[t]
    \centering
    \includegraphics[width=\textwidth]{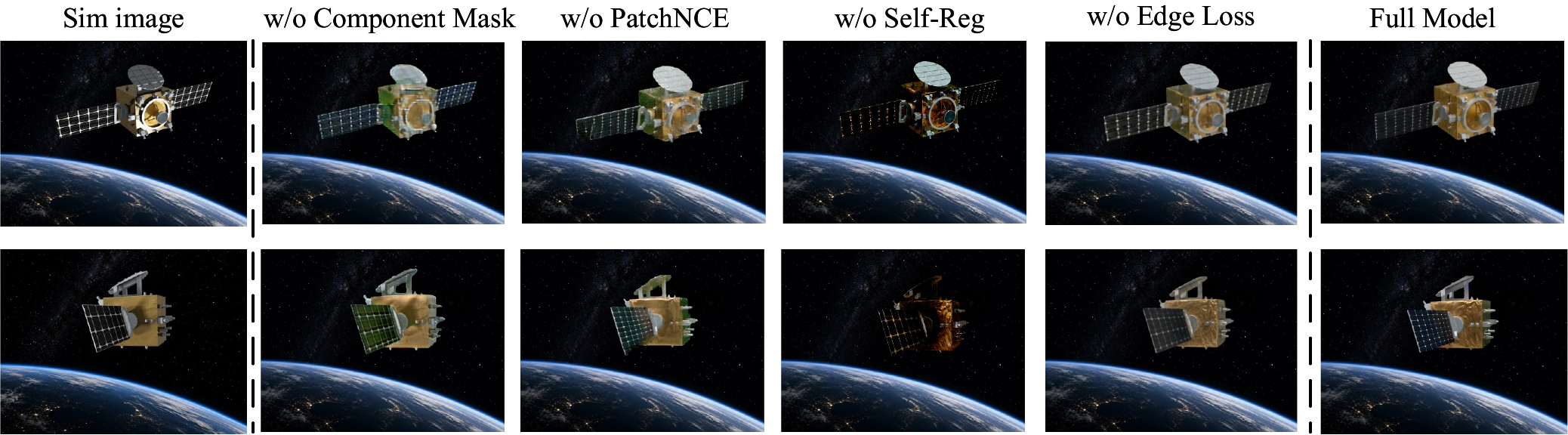}
    \caption{
    Visual ablation study of the proposed framework.
    Each column shows the result of removing one component from the full model.
    The full model better preserves satellite geometry, component layout, and structural boundaries while transferring realistic appearance.
    }
    \label{fig:ablation}
\end{figure*}

Fig.~\ref{fig:ablation} shows two representative visual examples. Removing component mask guidance weakens the semantic control over different satellite parts. As shown in the second column, the appearance of the solar panels and the main body becomes less clearly separated, and the translated texture tends to leak across component boundaries. Without PatchNCE loss, the global satellite pose is still preserved, but local details become less stable, especially around the solar-panel grid, the main body surface, and small protruding structures. Removing self-regularization introduces the most obvious appearance drift in the visualization: the satellite becomes overly dark and unstable, and the translated result deviates significantly from the input synthetic structure. Without edge-consistency loss, the overall appearance remains relatively reasonable, but the contours of the solar panels and the boundaries around the main body become softer and less distinct. In contrast, the full model produces more realistic component-specific appearance while better preserving the satellite pose, component layout, and structural boundaries.

\begin{table}[h]
\centering
\caption{
Ablation study of component guidance and structure-preserving losses.
FID measures image-level realism, while Mask IoU and Edge Error measure structural consistency.
}
\label{tab:ablation}
\begin{tabular}{lccc}
\hline
Variant & FID \(\downarrow\) & Mask IoU \(\uparrow\) & Edge Err. \(\downarrow\) \\
\hline
w/o Component Mask & 72.15 & 0.82 & 4.53 \\
w/o PatchNCE & 65.40 & 0.85 & 3.82 \\
w/o Self-Reg & 61.22 & 0.88 & 3.45 \\
w/o Edge Loss & 58.76 & 0.91 & 4.12 \\
Full Model & \textbf{54.32} & \textbf{0.95} & \textbf{2.15} \\
\hline
\end{tabular}
\end{table}

Tab.~\ref{tab:ablation} provides quantitative evidence consistent with the visual results. Removing component mask guidance causes the largest overall degradation, increasing FID from 54.32 to 72.15 and reducing Mask IoU from 0.95 to 0.82. This confirms that component-level semantic guidance is important for both realistic appearance transfer and structural consistency. Removing PatchNCE or self-regularization also degrades image fidelity and structural consistency, indicating that local contrastive consistency and pixel-level regularization help maintain the input structure during translation. When edge-consistency loss is removed, the Edge Error increases from 2.15 to 4.12, while the FID changes more moderately, suggesting that this loss mainly contributes to preserving sharp component boundaries and object contours. Overall, the full model achieves the best performance across all metrics, demonstrating the effectiveness of the proposed component-aware and structure-preserving design.

\section{Conclusions}
\label{sec:conclusions}

We proposed a component-level structure-preserving style transfer framework for satellite Sim2Real dataset construction. In the considered calibrated satellite setup, the method uses large-scale synthetic data and a limited number of unlabeled real observations to generate realistic satellite images while preserving the geometric annotations inherited from simulation. Component-level semantic guidance and mask-aligned style modulation are introduced to model part-specific appearance and reduce cross-component texture contamination.

An automated weakly paired data generation pipeline is further developed using ArUco-based camera-pose information, CAD-based rendering, and view consistency filtering, avoiding manual pixel-level registration. Structure-preserving losses, including PatchNCE, self-regularization, and edge-consistency loss, are used to maintain local structures and component boundaries during translation.

Experimental results show that, on this target-domain evaluation set, our method improves image-level fidelity and downstream satellite pose-estimation performance compared with representative baselines. Ablation studies confirm the importance of component mask guidance and structure-preserving constraints for maintaining image fidelity and structural consistency. Future work will investigate more robust component mask extraction, more diverse real-domain acquisition, additional satellite geometries, and evaluation under broader camera and illumination conditions.

\bibliographystyle{IEEEtran}
\bibliography{refs}

\begin{IEEEbiography}[{\includegraphics[width=1in,height=1.25in,clip,keepaspectratio]{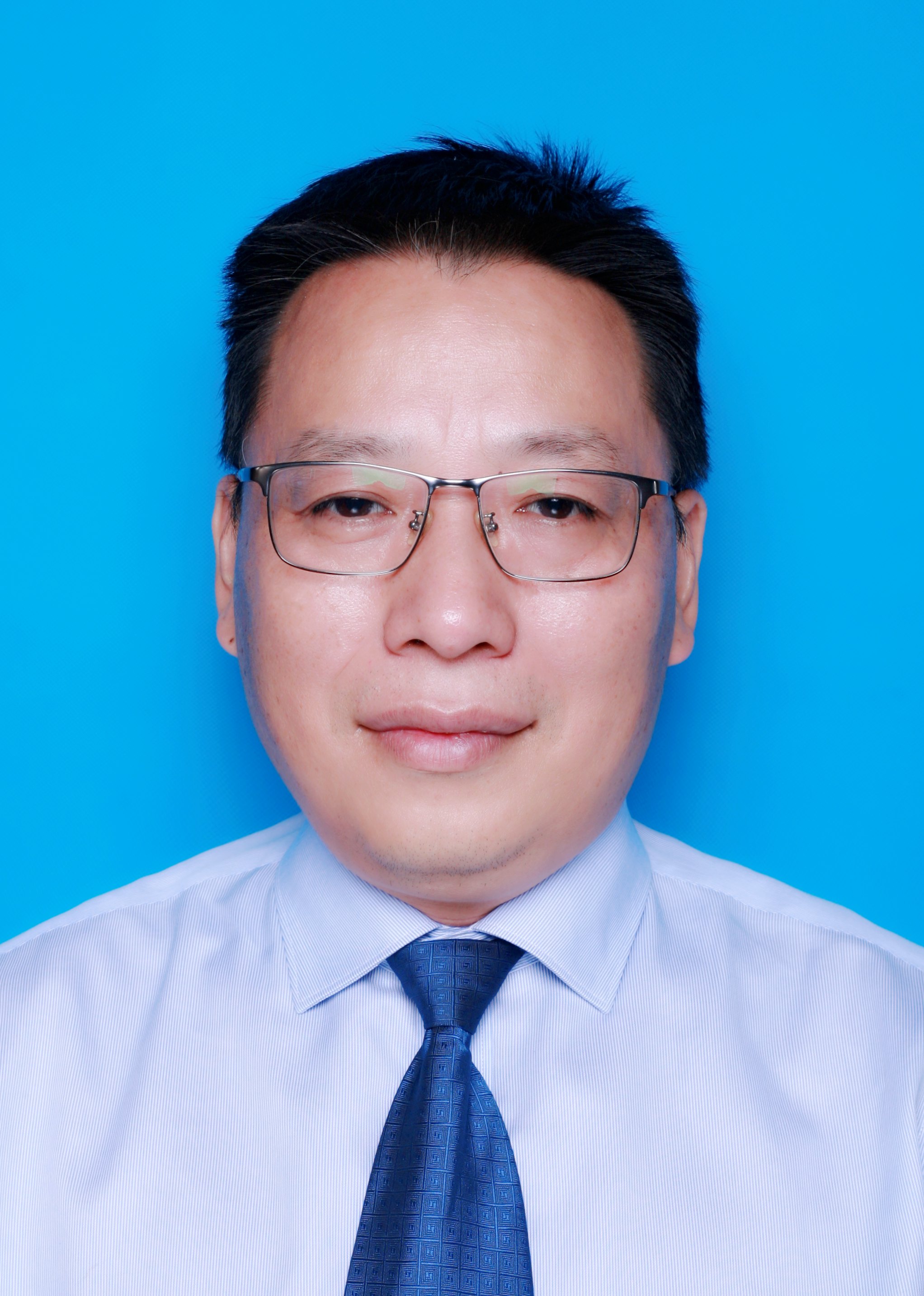}}]{Zongwu Xie}
Zongwu Xie received the B.S. degree in electrical engineering and automation from Harbin University of Science and Technology, Harbin, China, in 1996, and the M.S. and Ph.D. degrees in mechanical engineering from Harbin Institute of Technology, Harbin, in 2000 and 2003, respectively.
\end{IEEEbiography}

\begin{IEEEbiography}[{\includegraphics[width=1in,height=1.25in,clip,keepaspectratio]{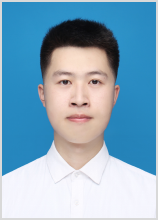}}]{Yonglong Zhang}
Yonglong Zhang received his Bachelor's degree in Mechatronic Engineering from Harbin Institute of Technology in 2024. He is currently pursuing the master's degree with the School of Mechatronics Engineering, Harbin Institute of Technology. His research interests include robot pose estimation and robotic arm manipulation.
\end{IEEEbiography}

\begin{IEEEbiography}[{\includegraphics[width=1in,height=1.25in,clip,keepaspectratio]{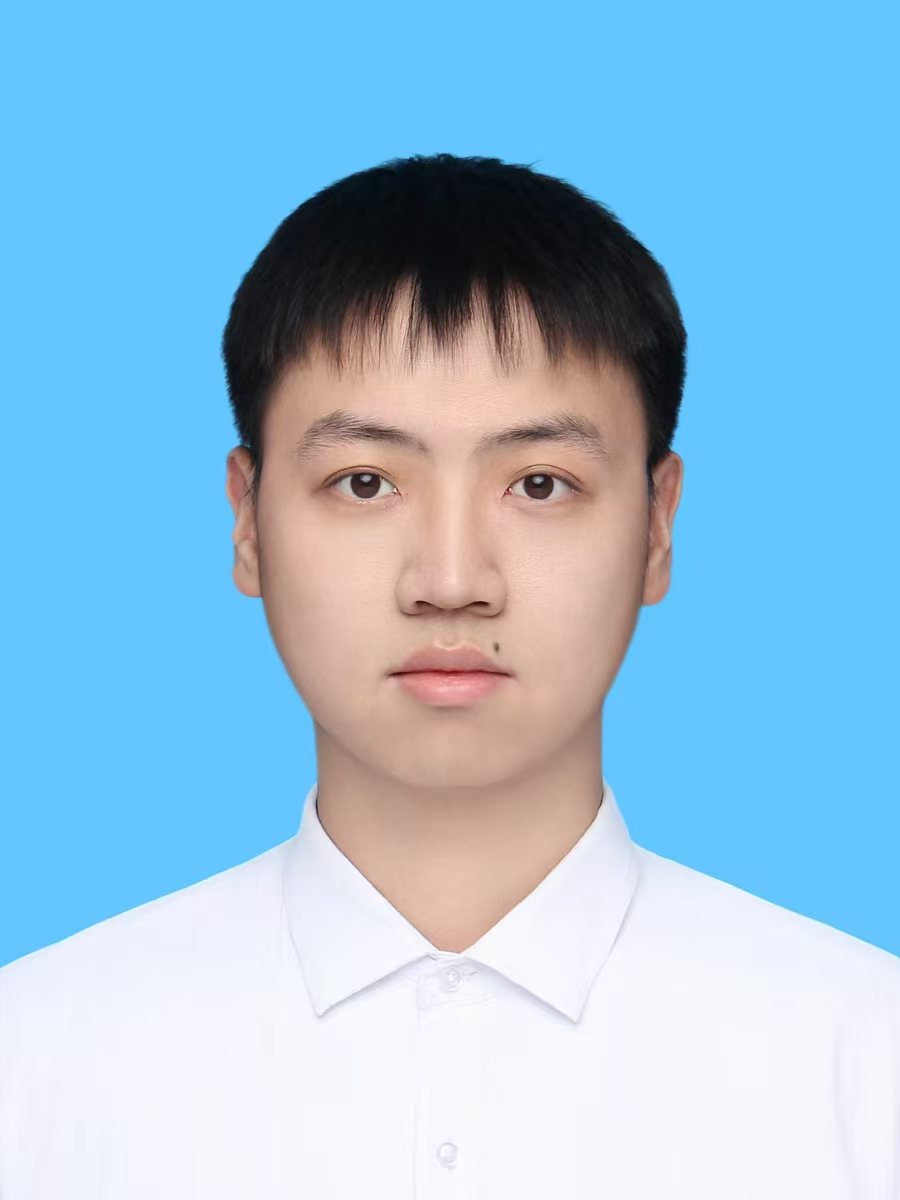}}]{Yifan Yang}
Yifan Yang received his Bachelor's degree in Mechanical Design, Manufacturing and Automation from Harbin Institute of Technology in 2025. He is currently a master's student (Class of 2025) at the School of Mechatronics Engineering, Harbin Institute of Technology. His main research focus is on robot pose estimation and arm manipulation.
\end{IEEEbiography}

\begin{IEEEbiography}[{\includegraphics[width=1in,height=1.25in,clip,keepaspectratio]{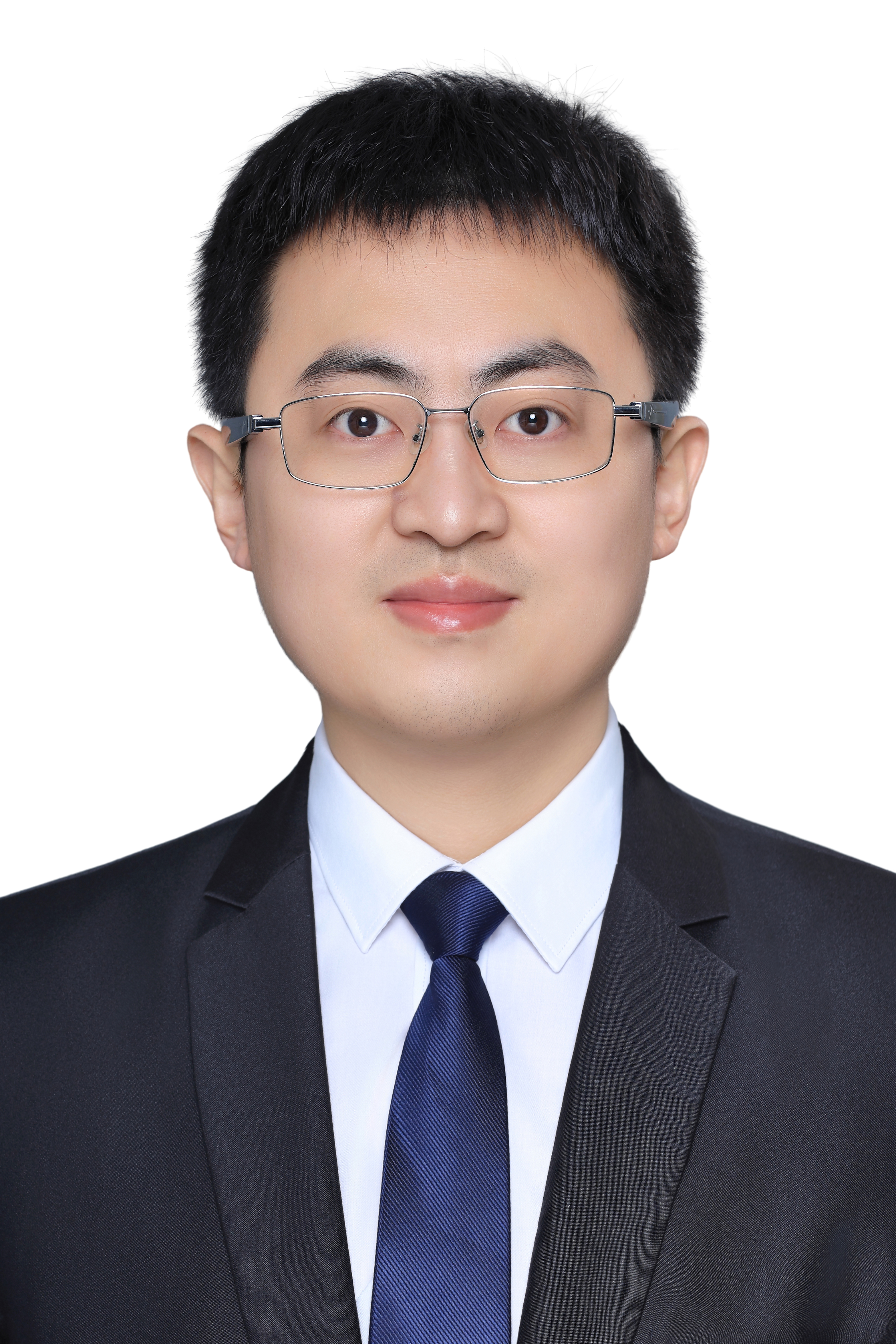}}]{Yang Liu}
Yang Liu was born in Handan, Hebei, China in 1990. He received the B.S., M.S. and Ph.D. degrees from the Harbin Institute of Technology, Harbin, in 2013, 2015 and 2020, respectively. Currently, he is an associate professor at Harbin Institute of Technology and works as the deputy secretary of the Youth League Committee. His research interests include space robotics, humanoid robot and computer vision. He has published 35 scientific papers, 16 invention patent and 2 textbooks. He has been selected for the Youth Talent Support Program and honored with titles including Heilongjiang Province Outstanding Youth.
\end{IEEEbiography}

\begin{IEEEbiography}[{\includegraphics[width=1in,height=1.25in,clip,keepaspectratio]{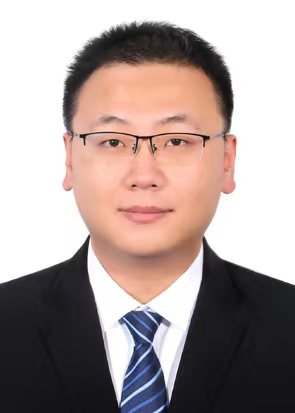}}]{Baoshi Cao}
Baoshi Cao received the B.S., M.S., and Ph.D. degrees in mechanical engineering from the Harbin Institute of Technology (HIT), Harbin, China, successively.
He is currently a Professor with the School of Mechanical Engineering, the State Key Laboratory of Robotics and Systems, HIT. His main research interests include space robots, humanoid robots, adaptive control, and trajectory planning of manipulators.
\end{IEEEbiography}

\begin{IEEEbiography}[{\includegraphics[width=1in,height=1.25in,clip,keepaspectratio]{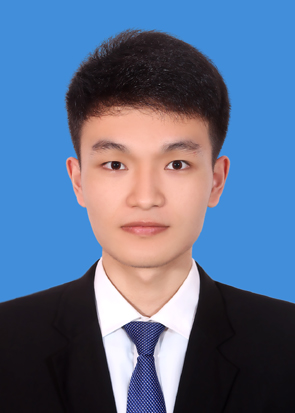}}]{Guanghu Xie}
Guanghu Xie received his Bachelor's degree in Mechatronic Engineering from Harbin Institute of Technology in 2020 and his Master's degree in Mechanical Engineering from Harbin Institute of Technology in 2022. He is currently pursuing the Ph.D. degree in mechanical engineering with Harbin Institute of Technology. His main research focus is on dexterous manipulation using robotic arms and hands.
\end{IEEEbiography}

\end{document}